\documentclass{article} 
\usepackage{iclr2027_conference,times}

\usepackage{amsmath,amsfonts,bm}

\def\eqref#1{equation~\ref{#1}}

\def\1{\bm{1}}

\DeclareMathAlphabet{\mathsfit}{\encodingdefault}{\sfdefault}{m}{sl}
\SetMathAlphabet{\mathsfit}{bold}{\encodingdefault}{\sfdefault}{bx}{n}

\usepackage{iclr2027_conference,times}
\usepackage[T1]{fontenc}
\usepackage{amsmath,amssymb,amsfonts}
\usepackage{booktabs}
\usepackage{tabularx}
\usepackage{array}
\usepackage{graphicx}
\usepackage{listings}
\usepackage{xcolor}
\usepackage{url}
\usepackage{microtype}
\usepackage{xcolor}
\usepackage{svg}
\usepackage{wrapfig}
\usepackage{float}

\usepackage{array}

\usepackage{longtable}
\usepackage{seqsplit}

\ifdefined\AppTableWidth\else
  \newlength{\AppTableWidth}
\fi

\title{Latent Assembly: Causal Discovery and Cross-Task Linking of Executable Operations in Language Models}

\author{Yuhan Chen, Yang Zhang \& Matthew Feng \\
University of California, San Diego\\
La Jolla, CA, USA \\
\texttt{\{chc248,yaz124,mafeng\}@ucsd.edu} \\
}

\iclrfinalcopy
\begin{document}

\maketitle
\lhead{Preprint}

\begin{abstract}
Can computations learned by language models be {extracted} as executable operations and assembled into new programs? We introduce a causal framework that recovers typed program states, discovers state-dependent operations and neural interfaces through which one operation’s output becomes another’s input. The framework identifies operations by their causal transformations rather than their physical locations and tests reuse through predicted intermediate states, causal execution, and recompilation.

In controlled transformers, our framework discovers ground-truth operation classes and composition laws starting from unlabeled neural activations. In pretrained Qwen, Gemma, and Llama models, we discover SELECT and LOAD operations connected by a reusable record-reference interface and validate their composition and recompilation. {Native-mediation experiments further show that the models' own computations use the consumer-interface subspaces recovered by our framework.}

Further, we show that three distinct operations discovered and qualified independently from separate tasks using our framework can be linked to execute semantically distinct tasks with no target-task fitting. We discover the frozen neural calling conventions through which the operations communicate. The three-operation chain achieves
{81\% overall accuracy against a 90\% model-native oracle. {Counterfactual changes to}
upstream arguments produce the downstream behavior implied by the altered
program. These results provide causal evidence that independently qualified computations can exchange executable internal
values through reusable interfaces.}

These results establish a
{scoped}
form of neural programmability
{on structured tasks}:
independently
{qualified}
computations can exchange arguments and execute together.
They open a path toward making computations acquired during training
accessible as components that can be inspected, tested, and deliberately
recombined.

\end{abstract}

\section{Introduction}
\label{sec:intro}

A compiled program can be taken apart. Given an executable, a decompiler recovers its functions, data structures, and calling conventions. A linker can then assemble separately compiled modules into a new program, provided their interfaces agree. Language models 
also generalize learned internal operations to solve tasks they were never explicitly trained on. Whether this computation can be taken
apart in an analogous sense---discovered as executable operations that can be lifted out of one task and run inside another---remains a central open question for mechanistic interpretability.

Mechanistic interpretability has uncovered several forms of reusable
computation. Circuits share components across tasks \citep{merullo2024circuit,he2025circuit}, function and task vectors causally invoke and compose behaviors \citep{todd2024function,lippl2025compositional}, and latent attention codes configure transformations reused on unseen compositions \citep{schug2025attention}.

These approaches motivate a further distinction between an operation and its
implementation or invocation. Distinct circuits can realize the same
computation \citep{bayatmakou2026distinct}; function vectors invoke behaviors,
while latent attention codes configure transformations.
{Likewise, passing decoded outputs between prompts can compose
behaviors without showing that one internal computation produces a state that
another computation natively consumes.} Establishing reusable executable
operations additionally requires discovering state-dependent transformation
laws and testing whether their outputs can be passed through qualified
interfaces to other computations.

To address these gaps, we introduce a \textbf{causal operation-discovery framework} that recovers typed program states, discovers state-dependent transformations, and qualifies their neural read/write interfaces. The framework tests candidate operations through causal execution, composition using their own predicted intermediate states, and recompilation of held-out programs.

We validate the framework in controlled transformers and pretrained
language models. In transformers compiled from known programs, it
recovers operation classes and composition laws, predicts held-out
compositions, and distinguishes native mediation from behaviorally
matched bypass routes. On structured tasks in Qwen, Gemma, and Llama,
it discovers \textsc{Select} and \textsc{Load} operations connected by a reusable record-reference interface and validates their composition and recompilation across model-specific physical implementations. {Native-mediation experiments in Qwen and Gemma further show
that the models' own computations use the consumer-interface subspaces
recovered by the framework.}

\begin{figure}[t]
    \centering
    \includegraphics[width=\linewidth]{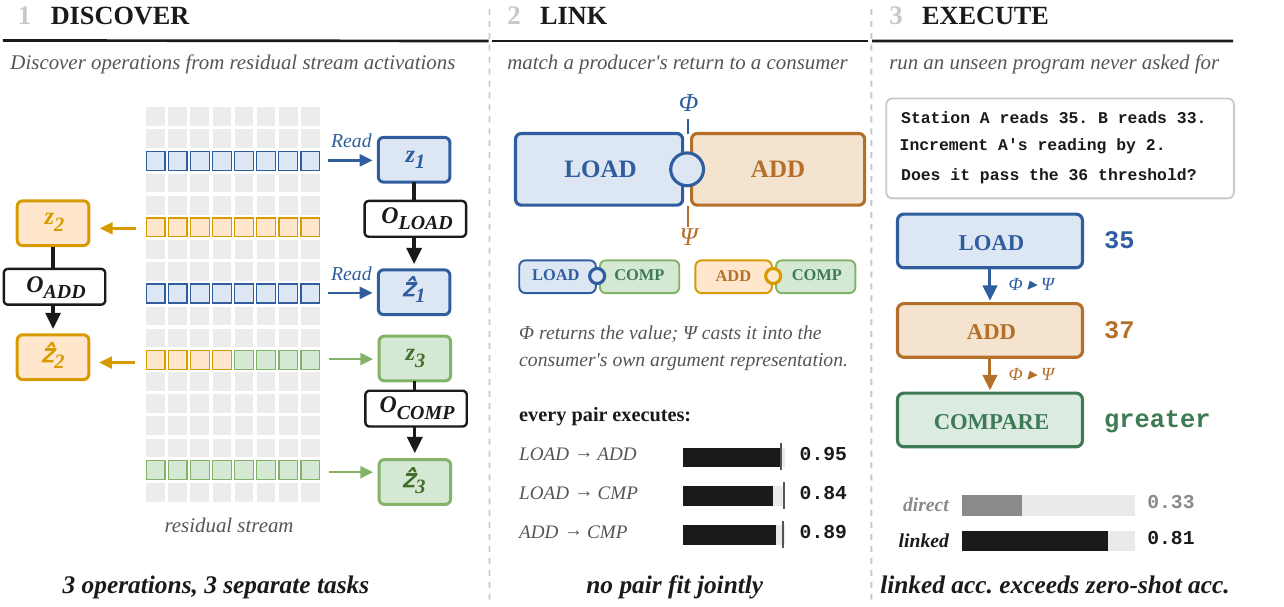}
    \caption{\textbf{Discover, link, execute.} Three operations independently discovered on
      separate source tasks and linked through frozen interfaces execute a held-out
      sensor-calibration program. (\S\ref{sec:prelim}, \S\ref{sec:linking}).}
    \label{fig:flagship}
\end{figure}

We then link \textsc{Load}, \textsc{Add}, and \textsc{Compare} operations independently discovered and qualified on separate source tasks in Qwen2.5-14B and Gemma-4-12B (Figure~\ref{fig:flagship}). Producer return maps and consumer casts form \textbf{neural calling conventions} that are fitted independently on source tasks and frozen for linking. The resulting library executes semantically distinct two- and three-operation programs without target-task operation fitting. A sensor-calibration chain reaches
{81\% overall accuracy and 100\% on strict-order comparisons.} Furthermore, changing an upstream argument produces the downstream behavior
implied by the altered program {in 99\% of counterfactual cases}.

Our core contributions are:
\begin{itemize}
    \item \textbf{A causal operation-discovery framework} that separates
    program states, transformation laws, and physical interfaces, with
    controlled and cross-model validation.

    \item \textbf{Neural calling conventions} that make independently
    discovered operations interoperable through separately qualified
    producer and consumer interfaces.

    \item \textbf{Cross-task execution of new two- and three-operation
    programs}, with counterfactual tests establishing that upstream
    arguments are transformed and consumed downstream.
\end{itemize}

\section{Related Work}
\label{sec:related}

\textbf{Interpreting computation.}
Mechanistic interpretability seeks algorithms implemented by neural networks
\citep{olah2020zoom,elhage2021mathematical}. Circuit methods localize mechanisms
\citep{olsson2022incontext,wang2023interpretability,conmy2023towards}, sparse
features decompose activations
\citep{bricken2023monosemanticity,cunningham2024sparse,ameisen2025circuit}, and
steering and function vectors causally control behavior
\citep{turner2023activation,zou2023representation,todd2024function}.
Our unit is the state-dependent operation and its executable interface.

\textbf{Causal abstraction and interfaces.}
Causal mediation and abstraction relate network behavior to high-level variables
under intervention \citep{vig2020investigating,geiger2021causal,geiger2025causal},
and distributed alignment search learns their neural realizations
\citep{geiger2024finding}. We {treat candidate operations and types as hypotheses and
causally test whether their predicted states can be consumed by independently
qualified downstream computations.} Concurrent hidden-API work \citep{ma2026hidden} identifies an
interface shared by multiple consumers. We additionally discover executable
transformations and link independently qualified producers and consumers.

\textbf{Validating interventions.}
Decodability need not imply use \citep{hewitt2019designing,belinkov2022probing,elazar2021amnesic},
localization need not predict editability \citep{hase2023does}, and subspace
patches can exploit dormant pathways \citep{makelov2024illusion}. Our
frame-changing interchange tests what a state carries, while compiled decoys
distinguish executability from native mediation.

{
\textbf{Reference representations and retrieval.}
Prior work identifies binding IDs, ordering-ID lookbacks, and
retrieval-conditioned rebinding circuits for maintaining and retrieving entity
references
\citep{feng2024entities,prakash2025languagemodelsuselookbacks,
oh-demberg-2026-retrieval}.
Our \textsc{Select}$\rightarrow$\textsc{Load} results extend this perspective
by testing whether a recovered reference acts as a typed value passed between
operations and supports composition and recompilation.}

\textbf{Program-level decompilation.}
RASP and Tracr provide explicit programs for transformers
\citep{weiss2021thinking,lindner2023tracr}. D-RASP reverses this direction,
reparameterizing trained transformers and causally extracting task-level
programs \citep{huang2026drasp}. We instead seek neural operations that remain
executable and reusable through qualified interfaces.

\section{Causal Operation Discovery}
\label{sec:prelim}

\subsection{Neural states and interventions}
\label{sec:prelim-states}

For residual stream $H^{(\ell)}(x)$, a checkpoint $c=(\ell,P)$ specifies a
layer boundary and token positions. Its substrate
$S_c(x)=\operatorname{Slice}(H^{(\ell)}(x),P)$ is the selected slice. The intervention
$\operatorname{do}(S_c=\widetilde S_c)$ replaces that slice and continues the
frozen model, leaving other coordinates unchanged at the intervention.

\subsection{Program states, operations, and interfaces}
\label{sec:prelim-ops}

\begin{wrapfigure}[17]{r}{0.46\textwidth}
    \vspace{-1.0\baselineskip}
    \centering
    \includegraphics[width=\linewidth]{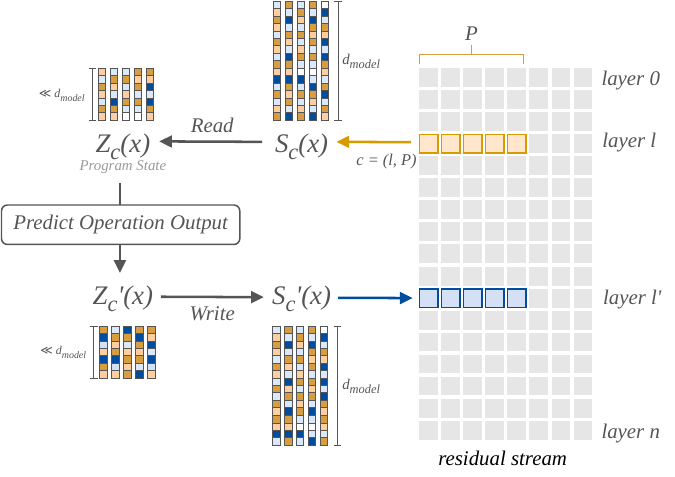}
    \vspace{-0.5\baselineskip}
    \caption{\textbf{Read, transform, write.}
    Compact program states connect residual-stream checkpoints.}
    \label{fig:read-transform-write}
    \vspace{-0.8\baselineskip}
\end{wrapfigure}

A program state $z$ {abstracts the information required to preserve the computation of interest.} Checkpoint-specific read and write maps connect it to neural states:
\begin{equation}
  R_c:\mathcal{S}_c\to\mathcal{Z}_c,
  \qquad
  W_c:\mathcal{Z}_c\times\mathcal{S}_c\to\mathcal{S}_c.
\end{equation}
{
Given a recipient substrate $S_c^r$, $W_c(z;S_c^r)$ installs $z$ without
replacing unrelated context. The write is causally sufficient if downstream
behavior follows.}

An \emph{occurrence} transforms states between two checkpoints. Let
$T_{\mathrm{in}:\mathrm{out}}^r$ denote the frozen continuation in recipient
context $r$, mapping an installed input substrate to the resulting output
substrate. A candidate law
$O:\mathcal{Z}_{\mathrm{in}}\to\mathcal{Z}_{\mathrm{out}}$ must satisfy
\begin{equation}
 R_{\mathrm{out}}\!\left(
 T_{\mathrm{in}:\mathrm{out}}^r\!\left[
 W_{\mathrm{in}}(z;S_{\mathrm{in}}^r)\right]\right)\approx O(z)
\end{equation}
on held-out installed states and recipient contexts, not only native runs.
Occurrences may lie in different layers, tasks, or models. After their
interfaces are independently qualified in common typed coordinates, we measure
agreement over matched states $z\sim\mathcal{D}$:
\begin{equation}
 d(O^{(i)},O^{(j)})=
 \mathbb{E}_{z\sim\mathcal{D}}\!\left[
 d_{\mathcal{Z}}\!\left(O^{(i)}(z),O^{(j)}(z)\right)\right],
\end{equation}
where $d_{\mathcal{Z}}$ is a task-specific output-state error. Agreement and
causal substitution qualify an \emph{operation class} over the tested domain.
Identity concerns the transformation, not its physical location; an operation
need not be primitive or have a unique decomposition.

\subsection{Behavioral fidelity and normalized causal efficacy}
\label{sec:prelim-fidelity}

Behavioral fidelity $F(a)$ is the fraction of held-out items on which
intervention arm $a$ produces the behavior implied by the installed program
state. Each test prospectively specifies a baseline $B$ without the causal
contribution under test and a reference fidelity $C>B$. We report
\begin{equation}
 \label{eq:rho}
 \rho(a;B,C)=\frac{F(a)-B}{C-B}.
\end{equation}
Thus $\rho=0$ matches the baseline and $\rho=1$ recovers the available
improvement. The reference usually installs a native target state through the
same interface. For incremental chain recovery, it is the downstream suffix
(Section~\ref{sec:linking-chain}). It is not a universal upper bound.

\subsection{Executing and reusing discovered operations}
\label{sec:prelim-exec}

\textbf{Execution.} A predicted state is executable when installing it through
$W_{\mathrm{out}}$ causes the downstream model to run the corresponding program.

\textbf{Composition.} For independently fitted operations with compatible
state spaces, we compute
$\widehat z_1=\widehat O_A(z_0)$ and
$\widehat z_2=\widehat O_B(\widehat z_1)$ on held-out programs.
\emph{Causal composition generalization} (CCG) requires executable states
generated recursively from predicted intermediates, without substituting
native ones. We report normalized efficacy $\rho$.

\textbf{Recompilation.} We assemble a held-out program from discovered
components and install it in a recipient that must execute it using its own
content. We report $F_{\mathrm{recomp}}$, since these tests do not specify
a separate normalization reference.

\textbf{Cross-task linking.} Independently discovered operations from different
tasks may have incompatible native representations. Separately qualified
producer return maps and consumer casts transfer their arguments
(Section~\ref{sec:linking}). No producer--consumer pairs or target-task
examples are used for fitting.

\subsection{The causal operation-discovery framework}
\label{sec:prelim-framework}

\textbf{(1) Recover interfaces.} Find candidate checkpoints and read/write
maps, qualify readability, writability, and downstream consumption.
\textbf{(2) Collect occurrences.} Gather pre- and post-states from designated
checkpoints.
\textbf{(3) Discover classes.} Fit compact state-dependent laws and test
held-out transfer. The \emph{shared-versus-separate} test compares a shared
law with occurrence-specific laws from the same estimator family; separate
fits can have more total parameters. A \emph{matched-input guard} requires
the same prediction for identical program states across occurrences,
excluding pooling that merely switches between context-specific rules.
\textbf{(4) Qualify operations.} Freeze candidate laws and test execution, CCG,
recompilation, and cross-task linking.
{\textbf{(5) Native mediation.} Lastly, we test if the model's
native computation uses our transplanted states.}

\section{Operation Discovery on Controlled Transformers}
\label{sec:controlled}

Section~\ref{sec:prelim} defines operations by their transformations, but
pretrained models provide no independent ground truth for a recovered
equivalence. We therefore begin with Tracr \citep{lindner2023tracr}, which
compiles RASP programs \citep{weiss2021thinking} into transformers with known
computational structure. {This setting lets us test three increasingly strong claims
against ground truth: transformation identity, predicted-state composition,
and native mediation. Operation labels are withheld during discovery and used
only for evaluation.}

\begin{figure}[t]
    \centering
    \includegraphics[width=\linewidth]{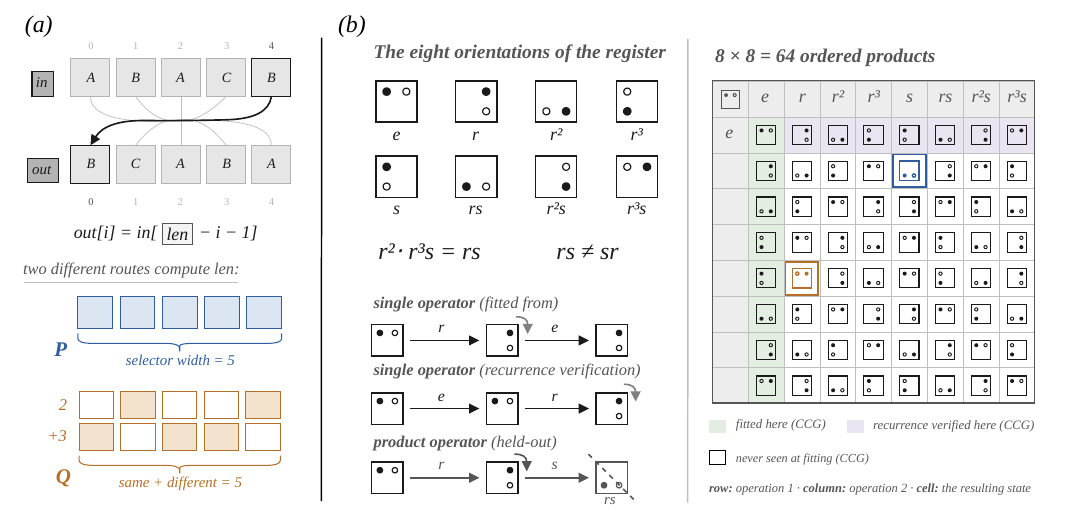}
    \caption{\textbf{Controlled task families.}
    \textbf{(a)} Sequence rearrangement (reverse). The operation studied is the routing rule. Distinct routes computing the same length operand.
    \textbf{(b)} Symmetry task. The eight states of $D_4$ and all 64 ordered products. Colors distinguish fitted atomic operations, recurrence checks, and
    held-out products.}
    \vspace{-2em}
    \label{fig:compiled}
\end{figure}

\paragraph{\textbf{Tasks.}}
Two compiled families test operation identity and composition
(Figure~\ref{fig:compiled}). \textbf{Sequence rearrangement} copies the token
at $j=f(i,\ell)$ to output position $i$, making the routing rule $f$ the
operation under study. Reversal, shift-by-two, and reflection recur with the
length operand computed through different RASP programs; shift-by-one,
shift-by-three, and doubling form singleton negative classes.
{The same operation therefore appears in distinct physical
routes, while similar implementations may realize different operations.}

\textbf{Symmetry tasks} maintain an 8-state register for the square's
orientation. The 8 elements of $D_4$, generated by quarter-rotation $r$
and reflection $s$, obey $r^4=e$, $s^2=e$, and $rs\neq sr$.
Single-operation organisms apply a symmetry followed by identity; product
organisms apply two symmetries, covering all $8\times8=64$ ordered pairs.
This tests if discovery recovers the order-sensitive composition law.

\textbf{Blind discovery.}
With operation identities withheld, we fit linear transition maps between the
states before and after each anonymous rearrangement occurrence. The 12
occurrences yield $\binom{12}{2}=66$ {pairwise
same/different decisions}. The shared-versus-separate test
(Section~\ref{sec:prelim-framework}) compares one shared map with two separate
maps, and the latter have more parameters. The matched-input guard additionally
excludes apparent sharing produced by branching on occurrence identity.

{All 66 decisions are correct, recovering all six ground-truth classes}. For same-operation pairs, sharing a transition
map adds no held-out prediction error ($0.0000$). For different-operation
pairs, the smallest added error is $0.2689$. An attention-head-overlap classifier is correct on only $33/66$ decisions, separating transformation identity from
physical implementation.

\textbf{Composition and execution.}
The recovered $D_4$ vocabulary assigns all 64 products to the correct class
at fidelity $1.000$. With twelve algebra entries withheld, all twelve are
predicted correctly, including the distinct outcomes of a noncommuting pair.

On a separate split, \textbf{causal composition generalization}
(CCG; Section~\ref{sec:prelim-exec}) tests whether operators learned only from
single-operation organisms recursively generate executable states. Each frozen
operator first transfers without refitting to the opposite program slot.
Across ten prospectively held-out compositions and all eight starting states,
predicted intermediate states, final states, and behavior are correct in all
$80/80$ cases. Substituting a wrong operation fails on all ten compositions;
reversing order succeeds on $0/6$ noncommuting and $4/4$ commuting pairs.
Native downstream consumers also correctly transform every installed
intermediate. One written state supports eight distinct subsequent outcomes,
ruling out a fixed final-answer interpretation.

\textbf{Native mediation.}
Executability alone does not establish native use. We compare behaviorally
identical transformers whose downstream computations consume either $z_2$
(\textsc{Mediated}) or a duplicate $\bar z_2=z_2$ during ordinary execution
(\textsc{Bypass}). A localized intervention on $z_2$ recovers $64/64$ target
transitions in \textsc{Mediated} and $0/64$ in \textsc{Bypass}, distinguishing their causal routes.

\section{Operation Discovery on Pretrained Models}
\label{sec:natural}

We next ask whether analogous structures can be recovered from pretrained
language models.

\label{sec:natural-states}

\begin{figure}[t]
    \centering
    \includegraphics[width=\linewidth]{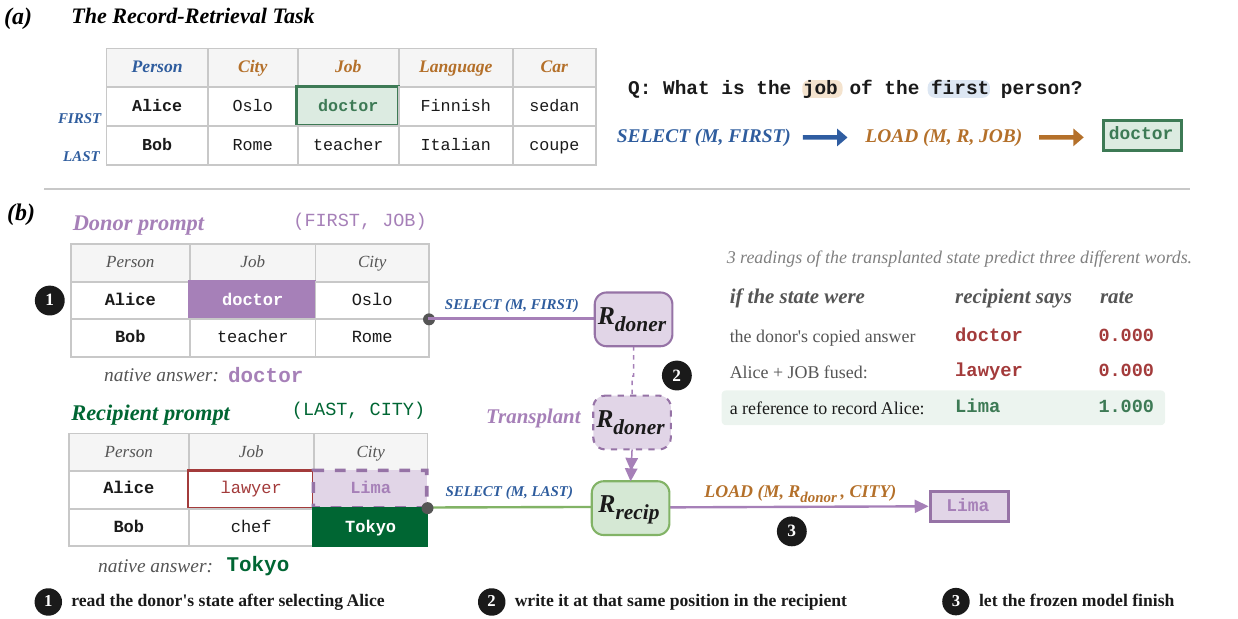}
    \caption{\textbf{Record retrieval and frame-changing interchange.}
    \textbf{(a)} Selecting the first record and loading its job returns
    \emph{doctor}. \textbf{(b)} We read the donor's intermediate after
    selecting Alice (1), install it in the recipient (2), and continue the
    frozen model (3). The recipient returns \emph{Lima}, Alice's city in its
    own table, rather than the donor's answer \emph{doctor} or the donor's
    field applied to Alice, \emph{lawyer}. This distinguishes a record
    reference from the donor's value or a record--field combination.}
    \vspace{-1em}
    \label{fig:donor-recipient}
\end{figure}

\textbf{Natural program family.}
We study record retrieval as a two-stage program:
\begin{equation}
    \operatorname{SELECT}(M,s)\rightarrow R,
    \qquad
    \operatorname{LOAD}(M,R,f)\rightarrow V,
\end{equation}
where $M$ is a record table, $s$ selects its first or last record, $R$ is the
selected record reference, $f$ specifies one of four fields, and $V$ is the
retrieved value. (Figure~\ref{fig:donor-recipient}a).

\textbf{Blind recovery of program state.}
The discovery procedure receives eight neural program states in anonymous
order, without semantic labels or a transition graph. Within the candidate
$2\times4$ shape, we exhaustively evaluate all $840=8!/(2!\,4!)$
arrangements, fitting $z_{ij}\approx\mu+a_i+b_j$ for each. The
\emph{interaction share} $I$ measures the fraction of state variation
unexplained by the two additive coordinate effects. Low $I$ means that changing
one coordinate has approximately the same neural effect across the other. On Qwen, the winning arrangement has $I=0.0353$, versus
$0.1651$ for the runner-up ($\Delta=0.1298$). Repeating the full search after
destroying cell identity gives a null 5th percentile of $0.1138$ and median
of $0.1577$. Only after freezing these analyses do we reveal the labels:
the recovered coordinates match \textsc{Selector}$\times$\textsc{Field}
in all $8/8$ cells.

\textbf{From program variables to typed operations.}
Factorization alone does not establish computational roles. We therefore
transplant the donor's intermediate after \textsc{Select} into a recipient
with different records and a different requested field
(Figure~\ref{fig:donor-recipient}b). Normally, the donor outputs Alice's job, while the recipient outputs Bob's city. Returning
\emph{Lima}, Alice's city in the recipient table, indicates that the
transplant changes the selected record while the recipient supplies the
field and its local value.

This outcome occurs at $1.000$ on Qwen2.5-14B, $1.000$ on Gemma-4-12B,
and $0.833$ on Llama-3.1-8B, with donor-field carryover of $0.000$ in every
case. The intermediate thus specifies \emph{which record to read}, supporting
the type $R:\mathrm{RecordRef}$ and the signatures
\begin{equation}
    \begin{aligned}
        \operatorname{SELECT}&:
        (\mathrm{Table},\mathrm{Selector})\to\mathrm{RecordRef},\\
        \operatorname{LOAD}&:
        (\mathrm{Table},\mathrm{RecordRef},\mathrm{Field})\to\mathrm{Value}.
    \end{aligned}
\end{equation}

These types distinguish the selected reference from the retrieved value and
inform the linking interfaces in Section~\ref{sec:linking}.

{
\textbf{Reference reuse beyond names and values.}
A reference map fitted only on unique-name city queries selects between
duplicate-name records at $0.912$ on Qwen and $0.963$ on Gemma, versus
approximately $0.50$ for name-only text. Once installed, the reference lets
the recipient's native \textsc{Load} retrieve jobs, numerical ages, and values
absent from fitting ($0.787$--$0.975$). Thus the reference identifies a record
without itself encoding the requested value.} 

\subsection{Discovering and qualifying SELECT and LOAD}
\label{sec:natural-selectload}

We study numerical and categorical retrieval, with all fields containing
numbers or categorical text, respectively. At the qualified interface, we fit
state-dependent laws on designated source examples: \textsc{Select} predicts
the displacement constructing $R$, and \textsc{Load} predicts the transition
retrieving the value addressed by $R$ and $f$. Their predicted states remain
executable on held-out occurrences. On Qwen2.5-14B, \textsc{Select} reaches
$1.000$ numerical and $0.846$ categorical causal execution, above positional,
shuffled, and matched-random controls. Numerical \textsc{Load} reaches
$1.000$ against a strongest control of $0.144$; categorical \textsc{Load}
reaches $0.856$ against $0.131$. Both branches independently localize
\textsc{Load} to $\mathrm{L}34\to\mathrm{L}42$.

{
\textbf{Transfer of the fitted \textsc{Load}.}
We next test the stricter question of whether the fitted value-producing
transition itself transfers to fields absent from fitting. We fit one map on
birth city, former job, age, and weekly hours, then evaluate current city,
current job, and commute time. The mixed-type map reaches $0.75$, $0.60$, and
$0.93$, respectively, against controls no higher than $0.20$; a
categorical-only fit raises current-city and current-job execution to $0.78$
and $0.89$.}

In the \textbf{CCG test} (Section~\ref{sec:prelim-exec}), \textsc{Load}
receives the reference predicted by discovered \textsc{Select}, rather than
the native intermediate. Normalized causal efficacy is $\rho=0.8182$ on
Qwen numerical retrieval and $0.7385$ on categorical retrieval under the
alignment-certified successor protocol (Table~\ref{tab:ladder}). For
\textbf{recompilation}, we reconstruct a reference in one context and install
it in another, requiring the recipient to execute the program using its own
content. Qwen reaches $F_{\mathrm{recomp}}=1.000$ in both domains, with
transported, reversed, and matched-random controls at zero.

\textbf{Replication across model families.}
Gemma-4-12B and Llama-3.1-8B also support discovered \textsc{Select} and
\textsc{Load} operations despite different layers, ranks, and causal timing.
CCG reaches $\rho=0.8667$ and $0.6289$, and recompilation reaches $0.9508$
and $0.6000$, respectively. Transported and reversed controls remain near
zero, although Gemma's matched-random control reaches $0.3115$.
Thus the framework recovers a common computational organization across
models with different physical realizations. Section~\ref{sec:linking}
tests whether independently discovered operations also support semantically
distinct tasks.

\begin{table}[!htb]
\centering
\caption{\textbf{Natural-operation qualification.} Num./cat. denote numerical
and categorical retrieval. CCG efficacy is $\rho=(F-B)/(C-B)$; $B$ and $C$
are the protocol's baseline and reference fidelities.{The final row reports the strongest transported,
reversed, or matched-random recompilation control.}}
\label{tab:ladder}
\smallskip
\small
\renewcommand{\arraystretch}{1.1}
\setlength{\tabcolsep}{4pt}
\begin{tabular*}{\linewidth}{@{\extracolsep{\fill}}lrrrr@{}}
\toprule
Measure & Qwen-num. & Qwen-cat. & Gemma-cat. & Llama-cat. \\
\midrule
\textsc{Select} execution & 1.0000 & 0.8462 & 0.8793 & 0.7083 \\
\textsc{Load} execution & 1.0000 & 0.8562 & 0.9375 & {0.9250} \\
\textsc{Load} margin over control & +0.8562 & +0.7250 & +0.8688 & +0.8938 \\
\midrule
CCG reference $C$ & 1.0000 & 1.0000 & 1.0000 & 0.9510 \\
CCG baseline $B$ & 0.9450 & 0.6750 & 0.1750 & 0.0000 \\
CCG fidelity $F$ & 0.9900 & 0.9150 & 0.8900 & 0.5980 \\
\textbf{CCG efficacy $\rho$} & \textbf{0.8182} & \textbf{0.7385} & \textbf{0.8667} & \textbf{0.6289} \\
\midrule
\textbf{Recompilation fidelity} & \textbf{1.0000} & \textbf{1.0000} & \textbf{0.9508} & \textbf{0.6000} \\
Strongest recompilation control & 0.0000 & 0.0000 & 0.3115 & 0.0500 \\
\bottomrule
\end{tabular*}
\end{table}

{
\textbf{Native mediation in pretrained models.}
We also apply Section~\ref{sec:controlled}'s mediation logic to test \textsc{LOAD} (as well as \textsc{ADD} and \textsc{COMPARE} introduced in Section~\ref{sec:linking}). Since our discovered operations write into a small subspace of model activations, we take pairs of prompts that differ in one place and have different answers. We give the model one prompt and swap in the other prompt's activation only within the subspace. In Qwen and Gemma, all three operations correctly switch to the other prompt when mediated $89\% - 99\%$ of the time. When we instead swap only the rest of the activations, switching occurs $0\% - 0.6\%$ of the time. This shows that pretrained models route through that subspace in its native computation.}

\section{Cross-Task Linking of Discovered Operations}
\label{sec:linking}

A producer may compute the right value in a representation another operation
cannot consume. We test whether \emph{neural calling conventions} make
independently discovered operations interoperable.

\textbf{Neural calling conventions.}
A checkpoint-independent argument space $\mathcal{V}$ separates the value passed
between operations from their native program states. Producer $A$ exposes its
result through a return map $\Phi_A$, and consumer $B$ incorporates that value
into its recipient state through a cast $\Psi_B$:
\begin{equation}
  \Phi_A:\mathcal{S}_{c_A^{\mathrm{out}}}\to\mathcal{V},
  \qquad
  \Psi_B:\mathcal{V}\times\mathcal{Z}_{c_B}\to\mathcal{Z}_{c_B}.
\end{equation}
For recipient substrate $S_B^r$, let $z_B^r=R_{c_B}(S_B^r)$. Linking constructs
\begin{equation}
  \widetilde z_B=\Psi_B\!\left(\Phi_A(S^A_{c_A^{\mathrm{out}}}),z_B^r\right),
  \qquad
  \widetilde S_B=W_{c_B}\!\left(\widetilde z_B;S_B^r\right).
\end{equation}
The write interface (Section~\ref{sec:prelim-states}) installs
$\widetilde S_B$ at the consumer's input checkpoint, after which the frozen
model continues.

\begin{figure}[H]
    \centering
    \includegraphics[width=\linewidth]{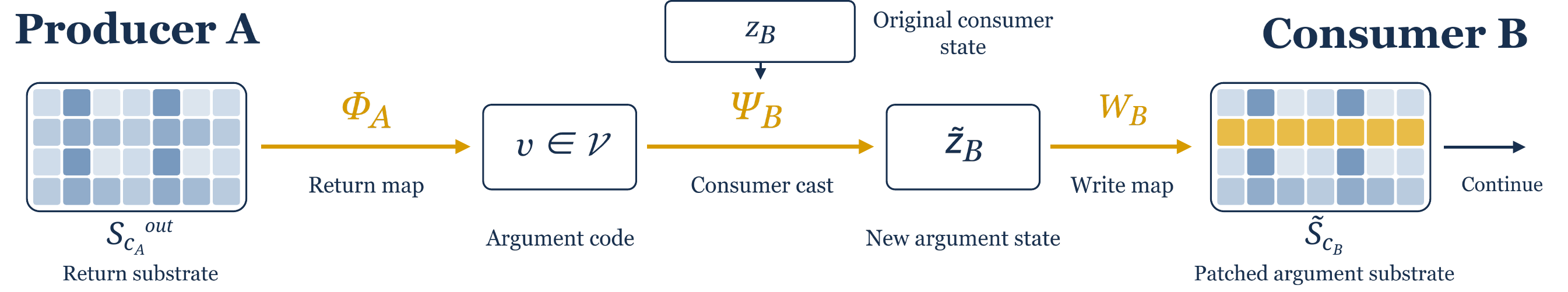}
    \caption{\textbf{Neural calling convention.}
    The return map $\Phi_A$ reads the producer's argument.
    The cast $\Psi_B$ combines it with the consumer's current state,
    and $W_{c_B}$ writes the result at the consumer checkpoint.}
    \label{fig:calling-convention}
\end{figure}

\subsection{Independently discovered operations interoperate}
\label{sec:linking-interop}

We discover and causally qualify \textsc{Load}, \textsc{Add}, and
\textsc{Compare} on separate record-retrieval, arithmetic, and
quantity-comparison tasks. Each producer's return map and each
consumer's cast are fitted independently on their source tasks, no producer-consumer pair fitted jointly. Installing reconstructed operand
states in held-out consumer examples tests if the interfaces transmit
executable arguments.

All three tested edges execute: \textsc{Load}$\to$\textsc{Add},
\textsc{Load}$\to$\textsc{Compare}, and \textsc{Add}$\to$\textsc{Compare}
reach fidelities of $0.950$, $0.840$, and {$0.890$}, against native-consumer
references of $0.970$, $1.000$, and {$0.940$}. Raw producer-state transplantation
yields $0.000$ on every edge. Changing the producer's result instead drives
the downstream computation implied by that value at $0.930$, $0.830$, and
{$0.800$}, respectively. Thus, the interfaces transfer specific arguments that
the consumer can compute with.

\subsection{Frozen links execute new two-operation tasks}
\label{sec:linking-frozen}

We fix source-qualified operations and interfaces before evaluating three
semantically distinct target tasks, with no target-task operation or interface
fitting. Rover routing requires retrieving and summing energy costs along a
two-leg route; production quota combines arithmetic with comparison; catalog
comparison combines retrieval with comparison. {Table~\ref{tab:links} reports each task against two baselines:
the model answering the target prompt in one pass (Direct), and a staged text
chain that runs the same decomposition with each intermediate result written
into the next prompt (Text).}

\begin{table}[!htb]
\centering
\caption{Frozen links on held-out tasks. $n$ denotes the number of held-out items. Each task uses its own primary
endpoint and efficacy normalization.}
\label{tab:links}
\smallskip
\footnotesize
\renewcommand{\arraystretch}{1.1}
\setlength{\tabcolsep}{3pt}
\begin{tabular*}{\linewidth}{@{\extracolsep{\fill}}llrccccr@{}}
\toprule
Task & Operations & $n$ & Linked & {Text} & Reference & Direct & $\rho_{\mathrm{link}}$ \\
\midrule
Rover routing & 2$\times$\textsc{Load}$\to$\textsc{Add} & 80 & 0.713 & {0.950} & 0.775 & 0.925 & 0.919 \\
Production quota & \textsc{Add}$\to$\textsc{Compare} & 180 & {0.828} & {0.972} & {0.844} & {0.578} & {0.981} \\
Catalog & 2$\times$\textsc{Load}$\to$\textsc{Compare} & {83} & {0.795} & {1.000} & {0.928} & 0.964 & {0.857} \\
\bottomrule
\end{tabular*}
\end{table}

{In a separate test, we install a retrieved price
directly at the price tokens of an otherwise unchanged Qwen catalog
prompt; the producer's reference price matches the price printed there.
The catalog comparison follows the retrieved price in all four
conditions for $79/96$ families ($0.823$), versus $0/96$ when the two
producer returns are exchanged and $78/96$ for a literal price edit.}

\subsection{Frozen links compose recursively}
\label{sec:linking-chain}

A longer program requires an argument to survive successive interfaces and an
intervening computation. We test this on sensor calibration: retrieve a station reading, apply its correction, and compare the result with an alert threshold. This requires a \textsc{Load}$\to$\textsc{Add}$\to$\textsc{Compare} program.

\begin{table}[!htb]
\centering
\caption{Sensor-calibration accuracy. {All results are reported in the form Qwen / Gemma.} The two-operation suffix executes \textsc{Add}$\to$\textsc{Compare} with the ground-truth upstream
\textsc{Load} return. The oracle supplies \textsc{Compare}'s native target state.}
\label{tab:chain}
\smallskip
\small
\renewcommand{\arraystretch}{1.1}
\begin{tabular*}{\linewidth}{@{\extracolsep{\fill}}lrrr@{}}
\toprule
Execution & Strict order ($n=100$) & Equality ($n=50$) & All ($n=150$) \\
\midrule
\textbf{Full linked chain} & {\textbf{1.000 / 0.720}} & {\textbf{0.420 / 1.000}} & {\textbf{0.807 / 0.813}} \\
Two-operation suffix & {0.990 / 0.720} & {0.520 / 0.980} & {0.833 / 0.807} \\
\textsc{Compare} oracle & 1.000 / {0.700} & 0.700 / {1.000} & 0.900 / {0.800} \\
\midrule
Direct & 0.490 / {0.390} & 0.020 / {0.180} & 0.333 / {0.320} \\
Chain-of-thought & 1.000 / {1.000} & 1.000 / {1.000} & 1.000 / {1.000} \\
{Text} & {1.000 / 0.710} & {1.000 / 1.000} & {1.000 / 0.807} \\
Best control (Noop / raw / matched random) & 0.470 / {0.460} & 0.000 / {0.000} & 0.313 / {0.307} \\
\midrule
\multicolumn{4}{@{}l}{\emph{Counterfactual tests}} \\
Wrong station $\rightarrow$ its implied relation & {0.990 / 0.900} & {1.000 / 0.960} & {0.993 / 0.920} \\
Wrong station $\rightarrow$ original target & {0.430 / 0.490} & {0.000 / 0.020} & {0.287 / 0.333} \\
No-op $\rightarrow$ recipient's base relation & {1.000 / 0.950} & {1.000 / 0.960} & {1.000 / 0.953} \\
\bottomrule
\end{tabular*}
\end{table}

{The Qwen chain exceeds direct answers by $0.473$ (95\% CI
$[0.39, 0.56]$), and the Gemma chain by $0.493$ ($[0.42, 0.57]$). In Qwen it
remains below chain-of-thought and the text-mediated chain (both $1.000$); in
Gemma it matches the text-mediated chain ($0.813$ vs.\ $0.807$). Staged text and CoT evaluates the same decomposition with decoded intermediate answers,  and they are not an alternative to what linking tests: whether one operation's internal result can be consumed by another.}

{
Furthermore, counterfactual tests verify that the chain uses both upstream operations.
Redirecting \textsc{Load} to another valid station produces the relation
implied by the altered program with $0.990$ accuracy. On a separate set of 143 strict-order items where this relation differs from both the original answer and the answer
obtained by skipping \textsc{Add}, the chain follows the altered program on
all $143/143$.}

{\textbf{Model-generated Decomposition.}
The preceding experiments specify the program externally. We now test whether Qwen can come up with the decomposition by itself. We use four different sensor-calibration subtasks, each requiring a different program over the \textsc{LOAD}, \textsc{ADD}, and \textsc{COMPARE} operations, and ask Qwen to output the program for each question. We decode greedily, take one sample, and do no retries or repair. Test questions use phrasings and station names that never appear in the examples. A plan may name only operations, stations, \textsc{CORRECTION}, and \textsc{THRESHOLD}, so a plan containing intermediate results or the final answer is rejected. One generic executor then runs whatever program the model wrote.}

{Qwen generates the correct program on $400/400$ questions, versus $98/400$ when the question is removed. A generic executor then runs the model-generated program, obtaining all four answers correctly in $0.99$ of families (95\% CI $[0.97,1.00]$). With each generated program
held fixed, edits to used inputs change the answer as implied on $0.988$ of
cases, while edits to unused inputs leave it unchanged on $0.995$.}

{This experiment shows manual front-end selection and wiring is not required for this task family. However, the runtime still places operands and invokes each operation separately, and we make no claim that the model internally executes the entire program in one forward pass.}

{
\textbf{Chain Scaling.}
With all interfaces frozen, we compose
$\textsc{Load}\to\textsc{Add}^{k}\to\textsc{Compare}$ programs up to ten
operations. Strict-order fidelity is $0.993$, $0.980$, $0.967$, $0.927$, and
$0.847$ at depths $2$, $3$, $4$, $6$, and $10$, while staged text remains
$0.993$--$1.000$. Given an exact input, each latent hop returns the exact value
on $0.86$--$0.94$ of items, so errors accumulate at an approximately constant
per-hop rate.}

\section{Discussion}
\label{sec:discussion}

Latent Assembly studies language model computation at the level of state-dependent transformations connected by typed causal interfaces. This abstraction separates what an operation computes from where it is implemented and how its arguments are represented. The recurrence of selection and retrieval across model families suggests that functional organization can remain recognizable despite differences in physical realization. Cross-task linking further makes this organization experimentally actionable: independently recovered computations can exchange intermediate values and execute programs beyond their source tasks. {Furthermore, our native-mediation experiments show evidence of native use of the transplanted interfaces on both controlled transformers and pretrained models.}

However, our experiments use structured tasks and a small operation inventory.
{Qwen can choose and connect the operations from a question,
but the runtime still invokes each operation separately. For each tasks involving \textsc{Add}$\to$\textsc{Compare} links, we found that the Qwen chains perform better on strict order items then equality items. This suggests that our discovered neural conventions can be imperfect. Also, text mediation remain strong execution baselines, particularly as programs grow longer and errors in latent arguments accumulate.} 

Nevertheless, our framework provides a concrete path toward broader settings: future work can discover candidate operations in richer tasks, link them through more comprehensive neural calling conventions, and test their reuse in more complex programs. Separating operation identity from physical implementation also enables investigation of which computational abstractions recur across models and how their interfaces differ. Furthermore, our counterfactual tests of operation executability and calling-convention reliability suggest a diagnostic strategy for failures of composition: intervening at intermediate stages can help distinguish failures of computation from failures of argument transfer. Ultimately, causal decompilation could make the computations acquired during training accessible through an explicit programming interface, enabling learned capabilities to be inspected, debugged, and deliberately recombined into new programs.

\section*{AI use statement}
Language models assisted with literature search and polishing the manuscript after human-written first drafts. Coding agents assisted with organizing the code repository for readability and presentation, {and with writing and debugging code during the rebuttal process.} The authors designed the experiments and checked all reported numbers against the experiment records.

\bibliography{iclr2027_conference}
\bibliographystyle{iclr2027_conference}


\end{document}